\documentclass[conference]{IEEEtran}
\IEEEoverridecommandlockouts

\usepackage{cite}
\usepackage{amsmath,amssymb,amsfonts}
\usepackage{algorithmic}
\usepackage{graphicx}
\usepackage{textcomp}
\usepackage[table]{xcolor}
\usepackage{booktabs}
\usepackage{url}
\def\BibTeX{{\rm B\kern-.05em{\sc i\kern-.025em b}\kern-.08em
    T\kern-.1667em\lower.7ex\hbox{E}\kern-.125emX}}

\newcommand*{\defeq}{\stackrel{\text{def}}{=}}
    
\begin{document}

\title{Can LLMs Generate Good Stories? Insights and Challenges from a Narrative Planning Perspective}

\author{
\IEEEauthorblockN{Yi Wang}
\IEEEauthorblockA{\textit{Autodesk Research} \\
San Francisco, United States \\
\texttt{ywang485@gmail.com}}
\and

\IEEEauthorblockN{Max Kreminski}
\IEEEauthorblockA{\textit{Midjourney} \\
San Francisco, United States \\
\texttt{maxkreminski@gmail.com}}
}

\maketitle

\begin{abstract}
Story generation has been a prominent application of Large Language Models (LLMs). However, understanding LLMs' ability to produce high-quality stories remains limited due to challenges in automatic evaluation methods and the high cost and subjectivity of manual evaluation. Computational narratology offers valuable insights into what constitutes a good story, which has been applied in the symbolic narrative planning approach to story generation. This work aims to deepen the understanding of LLMs' story generation capabilities by using them to solve narrative planning problems. We present a benchmark for evaluating LLMs on narrative planning based on literature examples, focusing on causal soundness, character intentionality, and dramatic conflict. Our experiments show that GPT-4 tier LLMs can generate causally sound stories at small scales, but planning with character intentionality and dramatic conflict remains challenging, requiring LLMs trained with reinforcement learning for complex reasoning. The results offer insights on the scale of stories that LLMs can generate while maintaining quality from different aspects.  Our findings also highlight interesting problem solving behaviors and shed lights on challenges and considerations for applying LLM narrative planning in game environments.
\end{abstract}


\section{Introduction}
Large language models (LLMs) are now being applied to a wide range of open-ended and creative tasks in and around games~\cite{GPTForGames,LLMsAndGames}. In particular, some recent games have begun using LLMs for open-ended story generation, demonstrating the potential for LLMs applied in this way to support novel forms of narrative interactivity~\cite{sun2023language,hua2020playing}.

Nevertheless, the capability of LLMs to produce high-quality narratives remains uncertain. Writing stories is difficult even for humans: authors typically need to establish a mental model of the story world, develop characters, mentally simulate different possible plotlines, curate compelling ones, and render them as text~\cite{kreminski2022unmet}. 
If LLMs are to produce good stories, it would seem that they must somehow perform the same tasks via autoregressive statistical sequence prediction---deviating from how human authors generally write~\cite{FlowerHayes}. 
Existing LLM-based story generation methods utilize either automatic metrics or human evaluation to assess the quality of generated stories\cite{yang2024makes}. However, automatic metrics typically address superficial aspects, such as lexical or semantic similarity to reference texts \cite{lin2004rouge,zhang2019bertscore}, while human evaluations are constrained by cost and the subjective nature of the evaluation criteria, making it difficult to achieve consistent results \cite{TTCW}.

On the other hand, what constitutes a good story is a central question for computational narratology---a field of study focused on modeling narrative structure using computational constructs, as well as creating and interpreting narratives with algorithmic processes\cite{mani2014computational}. Work in this field has led to the symbolic narrative planning approach to story generation \cite{TheStorySoFar, riedl2010narrative}, 
in which a desired world state is specified as the narrative goal, and event sequences leading to this goal---known as ``narrative plans''---are generated, subject to ``trajectory constraints'' that attempt to enforce narrativity~\cite{TrajectoryConstraints}. 

In this study, we aim to deepen understanding of LLMs' story generation capabilities by prompting them to generate narrative plans that adhere to constraints established in narrative planning literature. Specifically, we focus on \textbf{causal soundness} \cite{meehan1977tale}, \textbf{character intentionality} \cite{riedl2010narrative}, and the presence of \textbf{dramatic conflict} \cite{ware2013computational}. By evaluating the performance of LLMs in satisfying these constraints, we seek to gain insights into their ability to generate causally sound stories with believable character actions and dramatic conflicts.

For each of the three aforementioned aspects, we select a representative story domain from narrative planning literature. We first demonstrate that existing LLMs can solve the original version of the three examples with reasonable accuracy, thereby motivating the need for larger scale examples. We then create a benchmark consisting of parameterized variations of the examples, encoded via Answer Set Programming (ASP) \cite{gl8811988stable} to support automatic validation of the above three aspects. 

Our experiments show that GPT-4 tier LLMs can generate causally sound stories at small scales; however, planning with character intentionality and dramatic conflict remains challenging, requiring LLMs trained with reinforcement learning for complex reasoning. The results offer insights on the scale of stories that LLMs can generate while maintaining different kinds of quality.  Our findings also highlight interesting problem solving behaviors in connection with classical planning and shed light on challenges and considerations for applying LLM narrative planning in game environments, as well as the role of symbolic narrative planning in the LLM era.

Contributions of this study can be summarized as follows:
\begin{itemize}
    \item We show that existing LLMs can solve representative narrative planning tasks from literature, motivating the need for new benchmarking task sets with varied scales.
    \item We develop an LLM narrative planning benchmarking pipeline utilizing an Answer Set solver for causal transition simulation, plan validation and providing external feedbacks, via a translation of narrative planning problems into Answer Set Programs.
    \item We extend three well-known narrative planning tasks into parametrized task sets to facilitate evaluation of narrative planning involving causal soundness, character intentionality and conflicts.
    \item We evaluate LLMs' performance on the above task sets. Our findings provide insights on LLMs' capabilities in generating compelling stories, and shed lights on challenges and engineering practice and concerns for applying LLM narrative planning in actual game environments.
\end{itemize}

The taskset along with ASP encoding and LLM prompts used for all our experiments is publicly available at: \url{https://huggingface.co/datasets/ADSKAILab/LLM-narrative-planning-taskset}.

\begin{figure}
\centering
\includegraphics[width=0.48\textwidth]{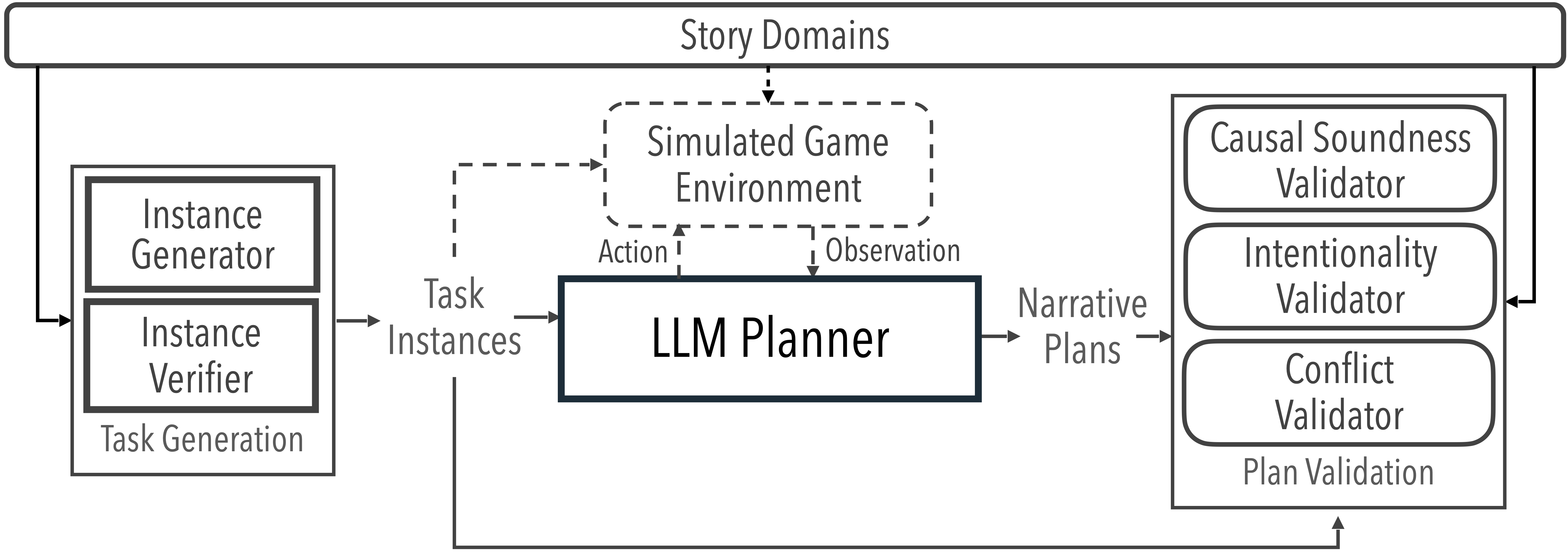}
\caption{Our evaluation pipeline consists of the task generation component, the plan validation component, and the LLM planner. Dashed lines represent processes only present in the external calibration mode. Rounded rectangles represent ASP-based components.}
\label{fig:pipeline}
\vspace{-10pt}
\end{figure}

\begin{figure*}[t]
\centering
\includegraphics[width=0.99\textwidth]{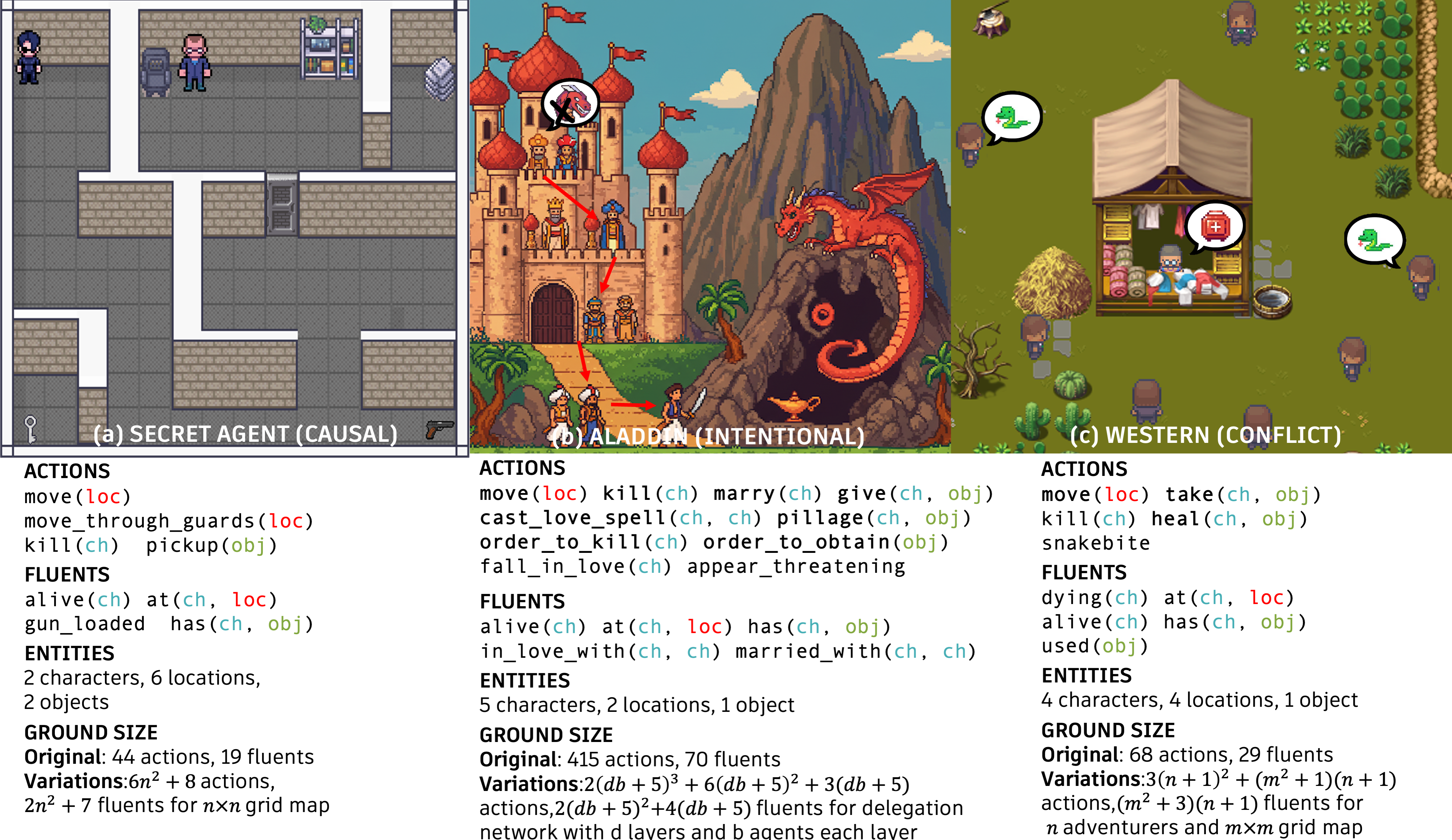}
\vspace{-5pt}
\caption{Description of the three story domains we use for benchmarking. Bold text indicates intentional actions. All the fluents are binary-valued except $has(ch, obj)$ in the Western domain takes a numerical value indicating the number of $obj$ that $ch$ has. The number of ground actions/fluents is calculated by counting how many ways to substitute each of $v_1, v_2, \dots, v_n$ in an action/fluents $p(v_1, v_2, \dots, v_n)$ - for actions it further take into account all possile characters performing the action. Action space size count includes nonexecutable actions. Ground symbol calculations exclude intention as part of the action symbols, despite being an argument of action predicates in the ASP encoding, because LLM-generated plans are not evaluated for correct intentions behind actions. Thus, intention symbols do not affect the search space size for LLMs.}
\label{fig:story_domains}
\vspace{-15pt}
\end{figure*}














\section{Related Work}

\subsection{Symbolic Narrative Planning for Story Generation}
Narrative planning methods~\cite{riedl2010narrative,TheStorySoFar} 
use symbolic search over spaces of possible character actions to generate plot structures. Planning often takes into account a wide range of narratological concepts, including character intention~\cite{riedl2010narrative}; character belief~\cite{wadsley2013belief} and meta-belief~\cite{Headspace}; conflict~\cite{ware2013computational}; surprise~\cite{Prevoyant}; tellability~\cite{Indexter}; and even interactivity~\cite{StoryAssembler}.
This makes the narrative planning literature a potentially rich vein of knowledge about the characteristics of a good story and how good stories can be computationally produced. However,
%
narrative planning domains are rigid and require specialized knowledge to author---leading some researchers and developers to pursue alternative means of story generation for games.

\subsection{Large Language Models for Story Generation}

Rapid progress in neural language modeling has brought about increased interest in text-oriented story generation, including via text continuation~\cite{fan2018hierarchical,TextContinuationRoemmele}; instruction-based prompting~\cite{Wordcraft}; more complicated prompt-chaining strategies, such as plan-and-write~\cite{PlanAndWrite}; emergent narrative via LLM-based character simulation~\cite{GenerativeAgents}; and even neurosymbolic approaches \cite{martin2018event}. Some of these neurosymbolic approaches combine LLMs with planning: TattleTale~\cite{TattleTale} uses a planner to generate plot structures for an LLM to follow; Kelly et al. \cite{ThereAndBackAgain} use an LLM to generate narrative planning domains which are then solved by a planner; StoryVerse \cite{StoryVerse} and WhatELSE \cite{lu2025whatelse} use LLMs for both top-down planning and bottom-up simulation; and Farrell and Ware \cite{farrell2024large} use an LLM as a heuristic to guide a symbolic narrative planner.

The coherence and prompt adherence of LLM-generated stories has improved considerably over the last few years, but these stories still exhibit noticeable deficiencies~\cite{TTCW,tian2024large,beguvs2024experimental}. How to address these problems is not entirely clear---in part because 
there is no clear consensus on how to evaluate stories~\cite{chhun2022human}, with even human experts using explicitly delimited rating criteria giving substantially divergent ratings to the same stories~\cite{TTCW}. By using narrative planning problems as concrete benchmarks for specific LLM story generation capabilities, we hope to build on existing attempts to operationalize narratological knowledge and better understand LLM storytelling strengths and weaknesses.


\subsection{Large Language Models for Reasoning and Planning}
Prior work has evaluated LLM ``reasoning'' via standard symbolic planning problems, broadly finding that LLMs have some ability to solve planning problems~\cite{IfWeTellThem} but tend to break down on larger or more complicated problems unless augmented by an ``LLM-modulo'' symbolic search framework~\cite{kambhampati2024llms}. LLMs might reasonably be expected to exhibit similar limitations on narrative planning problems, though it has also been found that LLMs seem to be able to leverage ``commmon-sense'' knowledge for improved planning ability on unobfuscated rather than obfuscated narrative planning problems~\cite{farrell2024large}. More research may help to illuminate when and how ``common-sense'' knowledge makes LLMs better at solving some planning problems than others.

\subsection{Answer Set Programming and Action Languages}
Answer set programming (ASP)~\cite{gl8811988stable} is a 
logic programming formalism that has been employed in many games-related domains, especially for 
procedural content generation~\cite{ASPforPCG}.
In particular, ASP has been applied to both planning~\cite{lifschitz2002answer} and narrative generation~\cite{chen2010rolemodel}, with some recent attempts drawing these approaches together~\cite{dabral2020generating} or using ASP to guide LLM-based story generation~\cite{Spleenwort}. One application of ASP is to model dynamic systems, in connection with {\em action languages}---formalisms for describing actions and their effects (PDDL is a well-known one). Many of these languages can be viewed as high-level notations of ASP programs structured to represent transition systems. The flexibility of ASP leads to various expressive possibility for actions languages, such as indirect effects, triggered actions, and additive fluents. In this study, we use the action language ${\cal BC}$+ \cite{babb2020action} to represent the story domains and execute through ASP solver {\sc Clingo}~\cite{Clingo}.
\section{Methodology}

\subsection{Preliminaries: Narrative Planning}
Though many works from the literature on narrative planning are built on top of the Partial Order Causal Link (POCL) planning framework \cite{williamson1996flaw}, for simplicity, in this study we prompt LLMs to generate total-order plans. We focus on total-order plans in the following review of preliminaries from symbolic narrative planning. 

A {\em story domain} is defined by
\begin{itemize}
\item $C$: a set of characters,
\item $V$: a set of fluents---each valuation of them is a {\em world state},
\item $A$: a set of actions each associated with preconditions and effects---preconditions define a subset of all possible world states in which the action is executable, while effects define how to update the world state once the action is executed.
\end{itemize}
In symbolic planning, fluents and actions are usually represented as logical predicates
such as  $location({\tt adam}, {\tt office})$.

\subsubsection{Causal Planning}
In the simplest case, narrative planning aims at generating causally sound sequences of character actions that lead to a set of target world states (called ``narrative goal''), which largely coincides with classical planning.

A {\em (causal) narrative planning problem} instance is a triple:
\vspace{-1mm}
\begin{equation}\nonumber
\langle \Pi, S_0, S_G \rangle
\vspace{-1mm}
\end{equation}
where $\Pi$ is a story domain. $S_0$ is the initial world state, and $S_G$ is a set of target world states specifying the narrative goal.

A solution of a (causal) narrative planning problem, called a {\em narrative plan}, is a sequence of actions $\langle a_0, a_1, \dots, a_n\rangle$ ($a_i\in A$), such that each $s_i$ satisfies the precondition of $a_i$, and $s_{n+1}\in S_G$, where $s_i (i>0)$ denotes the world state obtained from executing $a_{i-1}$ at state $s_{i-1}$, and $s_0 \defeq S_0$. 

In a narrative plan, we say that there is a {\em causal link} between an action $a_1$ and a following action $a_2$ by $p$, if executing $a_1$ makes some precondition $p$ of $a_2$ to be true (denoted $a_1 \stackrel{p}{\rightarrow} a_2$).

\subsubsection{Intentional Planning}
\label{sec:intentional_planning}
Riedl et al. \cite{riedl2010narrative} have extended this notion of narrative planning to go beyond classical causal planning by accounting for character goals. The work introduces the notion of character intentions, and requires a narrative plan to be decomposable to a set of ``frames of commitment'', in each of which a character takes actions driven by an intention.

A set of predicates is used to represent characters' intentions, of the form $intends(c, G)$, where $c$ is a character, and $G$ describes a subset of world states, representing that $c$ is taking actions towards leading the world state to $G$. Actions are divided into {\em intentional} and {\em unintentional} actions---the former can only occur when it's justified by an underlying intention.

A solution $S$ to a narrative planning problem is defined as {\em intentional}, if every occurrence of an intentional action is part of some commitment frame. A {\em commitment frame} is a subsequence $S'$ of $S$, with all actions performed by the same character $c$, with some intention state $intends(c, G)$ held throughout the course of this subsequence such that
\begin{itemize}
    \item the effect of the final action in $S'$ leads 
    to $G$,
    \item a motivating action preceding $S'$ has $intends(c, G)$ as its effect, and
    \item for each action $a_i (i < n)$ in $S'=\langle a_0, a_1, \dots, a_n\rangle$, there is a path of causal links from $a_i$ to $a_n$.
\end{itemize}

\subsubsection{Conflict Planning}
\label{sec:conflict_planning}
Ware et al. \cite{ware2013computational} further extended narrative planning to account for dramatic conflicts. Actions in a narrative plan are now further marked as {\em executed} or {\em nonexecuted} actions to allow characters to fail to achieve their goals, which further makes it possible to have conflict between commitment frames.

In a narrative plan, a causal link $a_1 \stackrel{p}{\rightarrow} a_2$ is said to be {\em threatened} by another action $a_0$ if

\begin{itemize}
    \item $a_0$ has effect $\neg p$;
    \item $a_0$ occurs after $a_1$, and before $a_2$.
\end{itemize} 
A {\em conflict} is further defined as a four-tuple $\langle c_1, c_2, a_1 \stackrel{p}{\rightarrow} a_2, a_0\rangle$, where:
\begin{itemize}
\item $c_1$ and $c_2$ are (possibly the same) characters;
\item there exist actions $a_0$, $a_1$ and $a_2$ in the plan, so that $a_1$ and $a_2$ form a causal link that is threatened by $a_0$;
\item $a_2$ belongs to a commitment frame of character $c_1$, and $a_0$ belongs to a commitment frame of character $c_2$;
\item at least one of $a_2$ and $a_0$ is an nonexecuted action.
\end{itemize}

\subsection{Benchmarking Pipeline}

We evaluate the capability of LLMs in solving narrative planning problems with three story domains (Figure \ref{fig:story_domains}), in terms of generating causally sound narrative plans, intentional narrative plans, and intentional narrative plans with conflicts. Our evaluation pipeline (Figure \ref{fig:pipeline}) leverage the Answer Set solver {\sc clingo} 5.7.1 as the formal verifier for causal soundness, intentionality and conflict, to automatically evaluate the narrative plans generated by LLMs.

In addition to {\em one-off prompting} to get narrative plans from LLMs, we assess the planning performance of an LLM-modulo approach \cite{kambhampati2024llms}, which incorporates {\em external calibration} for causal planning. This represents applications where LLMs are integrated in a game environment, allowing the environment to provide real-time feedback that guides and calibrates the LLM's output. Specifically, the LLM planner is paired with an ASP solver that simulates world state updates based on the narrative plan generated by the LLM, offering feedback on the causal soundness of the plan. The LLM is prompted to output one character action at a time, with the ASP solver executing this action within the simulated game environment. The LLM then observes the effects of the action and the updated world state before proceeding to the next action.

To scale up the original story domains for benchmarking, we create their parameterized variations. For each of the three story domains, an instance generator produces narrative planning tasks of varying scales by adjusting the introduced parameters. An instance verifier then removes unsolvable instances. The instance generator creates both ASP encoding for plan validation and natural language prompts for LLMs to generate narrative plans. The methods for parameterization, instance generation, and solvability verification are tailored specifically for each story domain. 

\subsection{Translating Narrative Planning Problems to ASP}

\subsubsection{Representing Causal Planning in ASP}

The causal planning problem involves identifying a sequence of transitions within a transition system, where the transitions correspond to the character actions updating the world state. PDDL solvers and POCL-based solvers are commonly used for causal planning in narrative contexts. In our study, we use a symbolic solver for plan simulation and validation rather than solving the planning problem itself. We consequently employ an ASP-based action language to represent transition systems due to its versatility in performing various inferences, support for functional arguments allowing flexible reification, and ability to extend for tracking character intentions and detecting conflicts. Following Babb and Lee \cite{babb2020action}, we represent the causal planning problem in ASP using action language ${\cal BC}$+.

Narrative plan validation is accomplished by encoding the plan as constraints within the main ASP program of the story domain, along with the ASP encoding of the initial world state and narrative goal constraints. The causal soundness of the narrative plan translates to the satisfiability of this combined ASP program. In the external calibration mode, we introduce a special predicate in the ASP program to provide external feedback (e.g., ${nonexec\_feedback}$("Destination isn't connected to starting location", $act({\tt secret\_agent}$, $move({\tt headquarter})$, $T$) with timestep $T$ of the error). After each LLM-generated action, these messages are included in the observation part of the prompt to guide the generation of the next action.

\subsubsection{Representing character intentionality}

Following Compilation 1 from \cite{haslum2012narrative}, we translate intentional narrative planning problems into classical causal planning problems. Actions are categorized into intentional and unintentional actions, with intentional action predicates including a special argument representing the intention behind the action. Preconditions for intentional actions require the corresponding intention to be held by the action's subject, and new predicates are introduced to track causal and motivational links. The occurrences of intentional actions require justification of their effects in terms of character intentionality, while each of their preconditions becomes justified for prior timesteps.

\subsubsection{Representing conflict}

Conflict detection is accomplished by introducing a new predicate to record actions that threaten causal links, tracking the conflicting commitment frames, the threatening action, and the effect of the prior action that is reverted. Actions are categorized into executed and nonexecuted---the latter do not change the world state but can justify the intentionality of previous actions. In the ASP encoding, rules are added to specify which effects of each action can potentially revert the effects of prior actions. For simplicity, nonexecuted actions are not required to be causally sound, although their intentionality must still be justified.


\subsection{LLM Prompts for Generating Narrative Plans}
Narrative plan generation prompts describe the following information in natural language:
\begin{enumerate}
    \item the list of characters, locations and objects, including description of their attributes and the relations between them that are static throughout all timesteps, including location connectivity,
    \item the narrative goal,
    \item the initial or current world state, for one-off prompting and external calibration mode respectively,
    \item the list of possible actions, including description of their preconditions and effects,
    \item for narrative planning involving intentionality and conflict, the definition of an intentional narrative plan (Sec. \ref{sec:intentional_planning}) in natural language,
    \item for narrative planning involving conflict, the definition of conflict (Sec. \ref{sec:conflict_planning}) in natural language.
\end{enumerate}
The LLM is prompted to generate a sequence of actions achieving the narrative goal (in one-off prompting mode) or a singular next action (in external calibration mode), with each action drawn exclusively from the provided list of possible actions. For tasks involving conflicts, the LLM is also prompted to specify whether each intentional action is executed. 

\subsection{Benchmarking Tasks}

We developed three sets of narrative planning tasks to comprehensively evaluate LLMs' narrative planning capabilities. These tasks are based on well-known examples from narrative planning literature, which we generalized into parameterized versions to enable procedural generation of variations.


\subsubsection{Secret Agent  \cite{riedl2004thesis} (Figure \ref{fig:story_domains} (a))}
In this story domain, a secret agent must navigate from the starting point to the mastermind's location, acquiring necessary items to pass through checkpoints and kill the mastermind. The domain features a single plannable character whose character goal coincides with the narrative goal. This example emphasizes classical causal planning, focusing on identifying causal dependencies between steps. The shortest plan for the agent to kill the mastermind consists of $7$  steps in our ASP encoding. We generate variations of the secret agent example by introducing random $n\times n$ grid maps, where obstacles, along with the gun and the key are randomly located on the grid. 

\subsubsection{Aladdin \cite{riedl2010narrative} (Figure \ref{fig:story_domains} (b))}
The Aladdin example, first introduced by Riedl and Young, demonstrates the concept of intentional narrative plans. In this domain, a dragon possesses a magical lamp, whose spirit can cast love spells. Only Aladdin can slay the dragon, but the king Jafar can command Aladdin to achieve specific goals. The task is to generate an intentional narrative plan where the king marries Princess Jasmine and the lamp spirit is killed. The challenge involves maintaining causal soundness, managing transmission of character intentions, and ensuring each intentional action fits within a commitment frame. For simplicity, unlike in the original IPOCL representation, where intention can be any arbitrary expression over world state, our ASP encoding allows only two types of intentions: $dead(Ch)$ (the intention to kill character $Ch$) and $in\_possession\_of(Ch, Obj)$ (the intention for character $Ch$ to possess object $Obj$). Only these goals can be delegated to another character. The shortest plan achieving Jafar's marriage to Jasmine and the lamp spirit's death consists of $12$ steps in our encoding. We generate variations of the Aladdin example by introducing intermediate delegation layers, controlled by two parameters: the number of delegation layers and the number of agents within each layer. A random tree structure represents the ``delegation network.'' Additionally, we constrain that only the knight Aladdin can move out of the castle, necessitating the task to find a path in the delegation network from the king to Aladdin.

\subsubsection{Western \cite{ware2013computational} (Figure \ref{fig:story_domains} (c))}

The story domain was introduced by Ware et al. to demonstrate planning for dramatic conflict. The story involves a parent attempting to steal medicine from a shop to save their child from a snakebite, while the sheriff tries to kill the parent for stealing, and the shop owner attempts to reclaim the medicine. The challenge lies in tracking (possibly failed) character plans and incorporating conflict into the narrative plan. The shortest plan leading to conflict consists of $6$ steps in our encoding. To scale up this domain for increasing difficulty of conflict occurrence, we randomly place $n$ newly introduced adventurer characters on a $m\times m$ grid, with the general store located at the center. Each adventurer can get snakebite at most once, and can legally obtain at most one medicine from the general store to heal themselves. The number of medicines is set to $n-1$, so that conflict only occurs when all the $n$ adventurers get snakebite, successfully navigate to the general store and attempt to obtain a medicine. For simplicity, the sheriff and parent characters are removed from these variations.
\section{Results}

\subsection{Overview}

\begin{table*}[t]
\centering
\caption{Success Rate of LLM Narrative Planning on the Original Secret Agent, Aladdin and Western Examples }
\label{tab:results_on_original_task_scale}
\begin{tabular}{|c|c|m{40pt}|m{40pt}|m{40pt}|m{40pt}|m{40pt}|m{40pt}|}
\hline
Domain                                                                                                                             & Condition                                                                                                             & \centering{gpt-3.5}  & \centering{gpt-4o}      & \centering{o1} & \centering{o1-mini}    & \centering{Claude-3.5} & Claude-3.7 \\ \hline
$\substack{\vspace{3mm}\\\text{\footnotesize Secret Agent}\\\text{(Causal)}}$           & One-off Prompting       &     $\substack{0/30\\ \text{$\sim$ 2400 tokens}, \\ \text{$\sim$ 5 sec}}$                                                                                             & $\substack{13/30 \\ \text{$\sim$ 900 tokens},\\ \text{$\sim$ 10 sec}}$ &  $\substack{30/30\\ \text{$\sim$ 2500 tokens},\\\text{$\sim$ 20 sec}}$    &     $\substack{28/30\\\text{$\sim$ 2,700 tokens}, \\ \text{$\sim$ 30 sec}}$ & $\substack{26/30 \\ \text{$\sim$ 850 tokens},\\ \text{$\sim$ 5 sec}}$ & $\substack{30/30 \\ \text{$\sim$ 2900 tokens},\\\text{ $\sim$ 30 sec}}$ \\ [8pt] \cline{2-8} 
                                                                                                                                   & External Calibration    &        \centering{$0/30$}                                                        & $\substack{29/30 \\ \text{$\sim$ 5000 tokens},\\ \text{$\sim$ 20 sec}}$ &  $\substack{30/30 \\ \text{$\sim$ 12,000 tokens}, \\ \text{$\sim$ 90 sec}}$    &  $\substack{30/30 \\ \text{$\sim$ 15,000 tokens,} \\ \text{$\sim$ 70 sec}}$    &        $\substack{30/30 \\ \text{$\sim$ 4500 tokens,} \\\text{$\sim$ 25 sec}}$        &    $\substack{30/30 \\ \text{$\sim$ 8900 tokens,} \\ \text{$\sim$ 80 sec}}$          \\ [8pt] \cline{2-8} 
\hline
$\substack{\vspace{1mm}\\\text{\footnotesize Aladdin}\\\text{(Intentional)}}$         & One-off Prompting      &  $\substack{0/30\\ \text{$\sim$ 2500 tokens,} \\ \text{$\sim$ 5 sec}}$                                                                                                         & $\substack{0/30 \\ \text{$\sim$ 2000 tokens,} \\ \text{$\sim$ 15 sec}}$  & $\substack{14/30 \\ \text{$\sim$ 10,000 tokens,} \\ \text{$\sim$ 100 sec}}$  &  $\substack{0/30 \\ \text{$\sim$ 7000 tokens,}\\ \text{$\sim$ 30 sec}}$  &  $\substack{0/30 \\ \text{$\sim$ 2200 tokens,} \\\text{$\sim$ 11 sec}}$  &   $\substack{15/30 \\ \text{$\sim$ 14,000 tokens,}\\ \text{$\sim$ 145 sec}}$               \\  [8pt]
\hline
$\substack{\vspace{1mm}\\\text{\footnotesize  Western}\\\text{(Conflict)}}$           & One-off Prompting       &   $\substack{0/30 \\ \text{$\sim$ 1500 tokens,} \\ \text{$\sim$ 5 sec}}$          & $\substack{3/30 \\ \text{$\sim$ 3000 tokens,}\\ \text{$\sim$ 20 sec}}$  & $\substack{20/30 \\ \text{$\sim$ 8000 tokens,} \\ \text{$\sim$ 70 sec}}$   & $\substack{6/30 \\ \text{$\sim$ 6000 tokens,}\\ \text{$\sim$ 30 sec}}$ & $\substack{2/30 \\ \text{$\sim$ 1650 tokens,} \\ \text{$\sim$ 8.5 sec}}$     &   $\substack{6/30 \\ \text{$\sim$ 15,000 tokens,} \\ \text{$\sim$ 150 sec}}$   \\ [8pt]
\hline
\end{tabular}
\vspace{-10pt}
\end{table*}

Table \ref{tab:results_on_original_task_scale} shows various LLM models' planning performance on the original secret agent, Aladdin and western examples. The success rates were computed based on 30 task-solving attempts. In addition to success rate, we also report rough scale of token count and time for solving the whole problem once. \footnote{Numbers for external calibration methods reflect only successful attempts.} As can be seen, the GPT o1 model is able to solve all the three examples, consistently on the secret agent example, and with around and above 50\% accuracy for the Aladdin and western example. GPT-4o has similar performance to Claude Sonnet 3.5, except that Claude-3.5 can solve secret agent with one-off prompting quite consistently. 
The reasoning models Claude Sonnet 3.7 and GPT o1 model have similar performance, except in that Claude-3.7 is significantly outperformed by o1 on the western example. GPT-3.5 was unable to solve any of the examples. All of the non-reasoning models struggle with the Aladdin and/or western examples, even the reasoning models o1-mini and Claude-3.7. The difficulty likely lies in understanding the complex computational definitions of intentionality and conflict, as well as failed actions, and applying them in problem solving without being misled by their prior understanding of these concepts. It can be seen that the token counts and solving time roughly indicate the difficulty of the problem for reasoning models, but not the case for the other models. 

\subsection{Detailed Analysis for Each Story Domain}

\subsubsection{Secret Agent (Causal Planning)}
\begin{figure*}[t]
\centering
\includegraphics[width=0.99\textwidth]{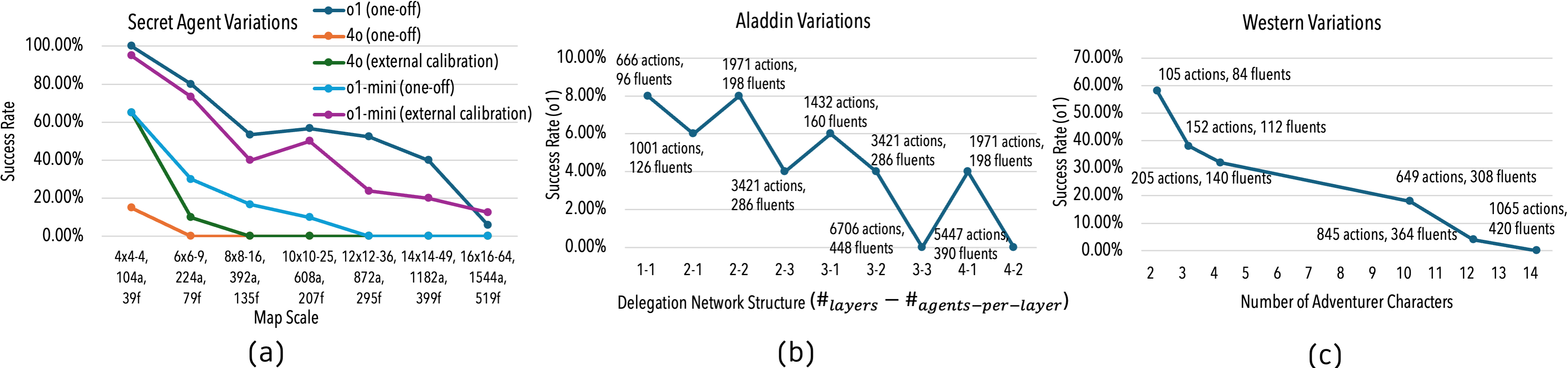}
\vspace{-10pt}
\caption{LLM Narrative Planning Performance on the Variations of the Three Story Domains}
\label{fig:variations_plots}
\vspace{-13pt}
\end{figure*}

Figure \ref{fig:variations_plots} (a) presents results for generated variations of the secret agent domain with different map scales to increase the difficulty of maintaining causal soundness. For each map scale, success rates are computed based on 50 solvable random configurations of obstacles, the key, and the gun. The o1 model, even with simple one-off prompting, achieved the best performance across all map scales up to 16x16. The performance of the o1-mini model with external calibration declined more slowly as map size increased, overall matching o1's performance. The o1-mini model without external calibration significantly outperformed GPT-4o with external calibration. For smaller scale instances, LLMs often found the shortest path, despite not being explicitly prompted to do so.

As the map scale increases, output plans tend to skip steps, leading to failure cases. These failed plans often maintain a correct high-level structure, such as finding the gun and key before passing the guard to kill the mastermind. For GPT-4o models with external calibration, sometimes the secret agent switches back and forth between objectives, such as moving toward the gun, then switching to obtain the key, and then returning for the gun, etc., resulting in unachieved goals. 

Overall, results from the secret agent domain indicate that post-GPT-4 LLMs can generally generate causally sound stories within certain scale. 
External calibration significantly enhances model performance, but also requires significant amount of additional time and tokens.


\subsubsection{Aladdin (Intentional Planning)}
Results from Table \ref{tab:results_on_original_task_scale} suggests that intentional narrative planning poses challenges except for certain reasoning models. Figure~\ref{fig:variations_plots} (b) presents the o1 model's performance on various sizes and shapes of delegation networks. For each parameter combination, 50 random configurations of edges representing ``loyal to'' relations were generated to benchmark the o1 model, with agents in lower layers loyal to those in upper layers. The network includes Aladdin as the lowest layer and the king as the highest, ensuring at least one path from Aladdin to the king. 

Comparing the success rate for an Aladdin instance and a secret agent instance of similar scale, e.g., Aladdin with $1-1$ delegation network and secret agent with $10\times 10$ map, the o1 model is able to achieve much better performance on secret agent. This highlights the fact that the difficulty of the Aladdin problem lies not only on the scale of the problem, but also the concept of intentional planning. 

Adding an additional delegation agent between Aladdin and the king significantly reduces the success rate. At the macro level, larger delegation networks result in lower success rates. At the micro level, network structure also affects success rates. A ``chain reporting structure'' without branching agents generally achieves higher success rates compared to other formations with the same number of agents.

For the o1 model, most failed plans are causally sound, but often with character actions lacking valid intention. For instance, Aladdin kills the dragon due to perceived threat but then obtains the lamp and casts a spell without valid motivation. Failed cases frequently introduce external motivations not established in the current plan, despite explicit prompts to avoid this. This suggests that the o1 model struggles with tracking character intentions under closed-world assumption.

\subsubsection{Western (Conflict Planning)}

Figure~\ref{fig:variations_plots} (c) displays the o1 model's performance on western variations with varying numbers of adventurers. Adventurers were randomly placed on a 5x5 grid, with the general store at the center, requiring 1-2 steps to reach it for each adventurers. This shifts the challenge from map navigation to tracking multiple character storylines. The o1 model handles up to 12 characters to plan for conflicts involving all characters. The success rate drops sharply from 2 to 3 adventurers (likely due to handling multi-party conflicts) and from 10 to 12 adventurers (likely due to exceeding context limits), with a slower decline in the middle range.

Comparing western instances to secret agent instances of similar scale, we find that conflict planning is more challenging than simple causal planning. The o1 model generally performs better on the western example than on the Aladdin example, likely due to the smaller action and state space in the western domain. The performance decline is slower in the western example, which is expected since the search space grows cubically with the number of characters in the Aladdin variations but only quadratically in the western variations.

However, even among instances with similar search space scales, the o1 model performs somewhat better on western instances compared to Aladdin instances. The o1 model's difficulties with the complex delegation structure and transmission of intentions in the Aladdin example suggest greater weaknesses around intentional planning than conflict planning. On the other hand, Claude-3.7 performs substantially better on the original Aladdin examples than on the western example (Table \ref{tab:results_on_original_task_scale}). Claude-3.7's weaker performance on the western domain seems to be due to difficulties adhering to the formal definition of an nonexecuted action as an action that \emph{would fail if attempted}, rather than an action that could be narrativized as failing for common-sense reasons.


In the original version of this example, we observe creative, technically correct plans leading to conflicts, though not aligned with common sense. For instance, one plan involves the parent taking medicine from the child in order to heal the child, conflicting with the child's intention to own the medicine to heal themselves. This suggests the o1 model can disregard its prior knowledge and operate strictly within the specifications of the symbolic planning domain.

As in the secret agent example, increasing the number of adventurers leads to plans that skip steps while maintaining reasonable high-level structure, with characters going to the general store or attempting to obtain medicine. We also observe interesting ``reward hacking'' behaviors. For instance, plans exploit the definition of conflict from \cite{ware2013computational}, where two nonexecuted actions can conflict, and our setting where preconditions of nonexecuted actions are not enforced. This results in agents merely ``thinking about'' taking medicine without execution to bypass map navigation. We observe other types of attempts to create shortcuts for conflict, such as an adventurer repeatedly getting snakebites or healing themselves to cause medicine shortage without involving other characters.

\section{Discussion and Future Work}

The secret agent example bridges our results with existing work on LLM for classic planning. Narrative planning, with constraints on character intentionality and conflict, is more challenging than classic planning. Unlike traditional planning tasks like blockworld and path planning, narrative planning may leverage LLMs' prior commonsense knowledge of story events and social interactions. However, our results suggest that the benefit from LLMs' prior knowledge, if any, does not counterbalance the inherent difficulty of narrative planning. Conversely, results from the Aladdin example indicate that prior knowledge may even have negative impact, misleading LLMs into invalid plans. This is evident in common failure cases where character actions are driven by motivations never established in the current plan.

This type of hallucination is intriguing. In the secret agent example, the o1 model was able to adopt a closed-world assumption for causal planning, rarely hallucinating nonexistent paths to the destination. However, the same model fails to adhere to this assumption for character intention.

A possible explanation for LLMs' failure at reasoning tasks is that pre-training and instruction-tuning corpora do not contain the true data-generating process~\cite{xiang2025towards}. We speculate that a similar reason underlies the failure in planning with character intentionality. Stories typically present sequences of character actions, by default without comprehensively specifying the intentions behind each action. Many fictions intentionally create ambiguity in character intentions to encourage multiple interpretations, making it difficult for LLMs to fully understand these aspects from textual data alone.

Despite the failure of other models, reasoning models show promise in narrative planning. However, symbolic planners such as Glaive \cite{ware2014glaive} remain superior for runtime narrative planning in games, considering both cost and running times (0.06 seconds for the original Aladdin and 28.41 seconds for the original Western). As of now, symbolic planners are still the preferred choice for efficient implementation. Symbolic planners will also likely remain superior at \emph{diverse} plan generation for the foreseeable future until the homogeneity of LLM output~\cite{beguvs2024experimental} is mitigated. 

Given the current success of connectionist approaches to AI, one might be inclined to believe that symbolic narrative planning will eventually be replaced altogether by data-driven approaches. There may be skepticism regarding our application of computational narratology to evaluate LLM-generated stories. The argument might be made that fuzzy, subjective, and complex concepts such as character believability and dramatic conflict---perhaps even the concept of ``a good story''---should be characterized statistically in the first place, which aligns with the operational principles of LLMs. Why, in this work, did we force LLMs to operate under clear-cut formal definitions of these concepts?

Broadly, we believe that even if symbolic narrative planning algorithms were to be superseded entirely, concepts and theories from computational narratology would remain valuable for human understanding of storytelling. They could aid in assessing story coherence and guide story generation toward specific standards, potentially through control mechanisms and evaluation metrics for LLM-based story generation. The main weakness of symbolic planning lies in the extensive engineering required to construct story domain specifications; future work might therefore improve the applicability of computational narratology by developing automatic methods to construct these specifications.

Our study has several limitations. We did not explore other narrative planning concepts, such as the computational characterization of character beliefs \cite{shirvani2017possible} and emotions \cite{shirvani2020formalization}. Additionally, we were unable to test a wider variety of LLMs or use larger sample sizes due to limited computational resources. The numbers of repeated experiments were a compromise between our computational budget and the stabilization of statistical results.  Time constraints also prevented the exploration of more sophisticated prompting schemes.




\bibliographystyle{IEEEtran}
\bibliography{bib}

\end{document}